# Tracking Control for Multi-Agent Systems Using Broadcast Signals Based on Positive Realness

Yasushi Amano, Tomohiko Jimbo, and Kenji Fujimoto

*Abstract*—Stochastic broadcast control is a control method for networked multi-agent systems. In this approach, each agent autonomously determines its own behavior (control input) via a probabilistic mechanism using only the signal sent from the central controller, without communicating with the other agents. Accordingly, it is useful for systems with numerous agents, or systems whose agents cannot communicate with each other. However, in engineering applications, it is difficult to manage the behavior of each agent. To solve this problem, this paper proposes a deterministic broadcast control method that makes a specified output signal follow the reference signal while letting the agents take different actions. Therefore, the agents make it possible to generate the autonomous behavior by deterministic mechanism. Since the differences of the behavior of the agents can be adjusted by selecting the design parameters, it is possible to assign their behavioral tendency in advance. Further, this paper proves the asymptotic stability of the proposed control system via hyperstability theory. When the tracking error with respect to the reference signal is large, all agents are active in control, and when the error is small, each agent autonomously decides whether to be inactive (stop) or to be active according to its own behavioral tendency, while ensuring that the overall system tracks the reference signal. Finally, numerical experiments are presented that verify the system can achieve the objective for various reference signal or environment.

*Index Terms*—Asymptotic stability, broadcast control, deterministic mechanism, passivity, tracking control

Y. Amano and T. Jimbo are with the Toyota Central R&D Labs. Inc., Nagakute, Japan (e-mails:e0738@mosk.tytlabs.co.jp; t-jmb@mosk.tytlabs.co.jp).

K. Fujimoto is with Graduate School of Engineering, Kyoto University, Kyoto, Japan (e-mail: k.fujimoto@ieee.org).

## I. INTRODUCTION

The internet of things (IoT) and cyber-physical systems [1],[2] enable high device–device functionality and efficiency through the exchange of information between devices via the Internet. Examples of such systems include energy systems such as power control systems in smart grids [3]–[6], collaborative systems involving multiple robots [7]–[10], multi-motor systems [11]–[13], and traffic systems such as traffic lights [14]–[16]. These are all multi-agent systems, in which it is difficult to realize centralized control by collecting and processing all the information in one place. Accordingly, it is necessary to design multi-agent systems that achieves goals by distributed control. Moreover, in distributed control, it is desirable to minimize the communication between agents to reduce the communication devices and data-processing system. It is also desirable for the agents to track any reference signal and to be robust against variation of the plant system and external disturbances.

There are a lot of research on distributed control of multi-agent systems. Some examples of distributed control are consensus control to match the state of each agent on the network [17],[18], coverage control to place agents in space to generate a specified distribution [19],[20], and distributed optimization in which the state variables of the agents are controlled to optimal values by only using on local information related to constraints and evaluation functions [21],[22]. In such methods, only neighboring agents communicate with each other. Accordingly, communication devices are required, whereas such devices would increase the cost.

To overcome such problems, broadcast controllers have been proposed [23]-[26]. In this approach, the controllers achieve tracking control by using the measurement of the tracking error only without communication among agents. This approach is effective for systems with numerous agents or systems in which communication between agents is difficult. Moreover, in broadcast control, each agent has the same local controller and determines the control input value autonomously by using the same broadcast signal that is sent simultaneously to all the agents. The behavior of each agent depends on its initial state, which limits achievable desired behavior. To avoid this problem, some methods using stochastic local controllers have been proposed. Ueda et al. [23] reported that the control objective can be achieved using stochastic control of the ON–OFF contraction movement of a muscle, by only using the measurement of the difference between the desired length and the current length of the muscle. Azuma et al. [24], [25] and Ito et al. [26] have also reported that problems such as coverage control can be handled using stochastic broadcast control.

The broadcast control is effective for the system with numerous agents. However, engineering applications do not always have numerous agents. For example, an energy management system has many power sources agents and a multi-motor system has some motor agents. In each of these situations, it is easy to design a local controller based on the characteristics of each agent. However, for agents that operate stochastically, it is difficult to design the local controller since its behavior is stochastic.

When the local controller can be design as described above, the following becomes possible. For an energy management problem in which the demand power is generated by multiple power sources, it is desired to use all the power sources when the required power is large and to stop the inefficient power sources when the required power is small. When generating power using a generator (internal combustion engine, etc.), the efficiency generally decreases if the power supply is low owing to the effects of mechanical losses such as friction. Therefore, it is beneficial in terms of efficiency to stop unnecessary power sources. In addition, in a multi-motor system, it is generally inefficient to generate small torque in the motor to achieve the purpose. Consequently, it is desired that motors specified beforehand (for example, according to the efficiency) stop and remaining motors produce the desired torque cooperatively. In other words, it is useful for each agent to divide the roles of starting, stopping, and output adjustment automatically. The roles of agents can be preset by the design parameters.

To ensure the reliability of multi-agent systems, it is necessary to prove their stability, especially in a target tracking problem, the asymptotic stability to assure that the output of the system tracks the reference signal. As stable analysis methods applicable to multi-agent systems, previous studies [27]-[29] have proposed methods based on passivity. The passivity is an effective tool to consider the stability of a multi-agent system. For example, parallel coupling of passive systems also satisfies passivity. Furthermore those methods are applicable to nonlinear systems as well. Such a passivity based approach only proves input-output stability of the closed loop system but it does not ensure asymptotic stability.

In contrast, in real-life environments, living organisms realize distributed control-based automatic division of roles, which is the aim of the present study. A representative example of such control is gait control in multi-legged animals, and some studies are being conducted to elucidate the underlying control principle. Owaki et al. [30],[31] previously proposed autonomous distributed control methods to realize collaboration among the legs of four-legged robots. Such controls can generate gait patterns, that is, realize division of roles for the legs, such as which leg is the support leg and which is free, by using only the information about each leg. Moreover, it has also been stated that such control can generate various gait patterns depending on the walking speed. That is, these approaches can automatically generate the division of roles for the legs depending on the walking speed. However, those studies focus exclusively on reproducing the principles of gait in living organisms and applying them to control. Thus, system stability was not investigated.

This paper proposes a multi-agent system with a deterministic decentralized controller that achieves to track the reference signal. The proposed system has the following features:

1) Each agent uses no information of other agents.
2) Each agent decides its action using the broadcast signal only.
3) The behavioral tendency of each agent can be designed by control parameters.

Furthermore, we analyze the stability of the proposed control system and clarified the conditions under which the tracking error converges to zero. The proposed control method achieves that a single output signal follows the reference using the control inputs generated by multiple agents. For example, a multi-motor system to control the torque of one output shaft or an energy management system to realize the desired power with multiple power sources. The proposed agent consists of a switching variable, which is an integral of the tracking error, and a switching function, which defines the control input based on the switching variable. This approach enables the agents to produce different outputs, even if all agents use the same broadcast signal. Moreover, a sufficient condition for the tracking error to converge to zero is given. Further, it is possible to design a division of roles that completely stops some agent when they are not necessary . In addition, we propose a method of producing a continuous control input to reduce the oscillations generated by switching multiple control inputs.

The remainder of this paper is organized as follows. Section Ⅱ formulates the problem. Section Ⅲ presents the details of the proposed method and the proof of stability. Section Ⅳ outlines the numerical experiments conducted and analyzes the results obtained. Section V discusses the effectiveness of the proposed method and its issues. Section VI presents concluding remarks.

## II. Problem Formulation

This section describes the tracking control system and the purpose of the control used in this study.

### A. Controlled System

In the tracking control system, as shown in Fig. 1, $G_p$ is the plant system and $G_c$ is the controller consisting of the multi-agent system. The $G_p$ can be expressed as follows:
$$\dot{x}_p = A_p x_p + b_p u_p \\ y_p = c_p x_p + d_p u_p \qquad (1)$$
and is assumed to satisfy Assumption 1 below. Here, $u_p \in \mathbb{R}$ is the input, $y_p \in \mathbb{R}$ is the output, $x_p \in \mathbb{R}^n$ is the state variable, and $A_p \in \mathbb{R}^{n \times n}$, $b_p \in \mathbb{R}^n$, $c_p \in \mathbb{R}^{1 \times n}$, and $d_p \in \mathbb{R}$ are constant matrices and vectors.

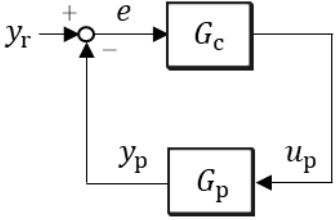

Fig. 1. Controlled system comprising $G_c$ and $G_p$.

Assumption 1: Equation (1) is a linear single-input, single-output strictly positive real system.

Note that a single-input single-output system as in Equation (1) is said to be positive real if there exist positive definite symmetric matrices $Q \in \mathbb{R}^{n \times n}$ and $P \in \mathbb{R}^{n \times n}$, and $l \in \mathbb{R}^{1 \times n}$, $w \in \mathbb{R}$ satisfying the following conditions:
$$A_p^T P + P A_p = -Q - l^T l \\ P b_p = c_p^T - l^T w \qquad (2) \\ 2 d_p = w^2.$$

Moreover, the multi-agent system $G_c$ is configured as shown in Fig. 2, where the input $e \triangleq y_r - y_p$ is the broadcast signal and $y_p$ is the output and $y_r \in \mathbb{R}$ ($y_r < \infty$) is the reference signal, and the control input $u_p$ is set as the sum of $u_{pi} \in \mathbb{R}$ generated by agents $g_{ci}$ ($i = 1, 2, 3, \ldots, m$) as follows:
$$u_p = \sum_{i=1}^{m} u_{pi}. \qquad (3)$$

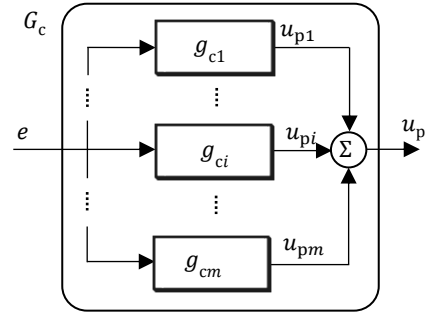

Fig. 2. Controller $G_c$.

### B. Control Objective

In this study, the output $y_p$ of the plant system tracks its reference signal $y_r$. That is, the control problem is to design $g_{ci}$'s so that the following $e$ becomes to zero.
$$e \triangleq y_r - y_p. \qquad (4)$$
Here, the following assumption is made regarding $y_r$.

Assumption 2: $y_r$ is assumed to be the output of
$$\dot{x}_r = A_p x_r + b_p u_r \\ y_r = c_p x_r + d_p u_r \qquad (5)$$
where $u_r$ is a constant and $y_r$ may be time-varying (not necessarily constant ).

To be able to track the above reference signal $y_r$, it is assumed that the control input satisfies the following conditions.

Assumption 3: $U_{\min} \leq u_r \leq U_{\max}$, where $U_{\min} \triangleq \sum_{i=1}^{m} U_{\min}^i$, $U_{\max} \triangleq \sum_{i=1}^{m} U_{\max}^i$, $U_{\min}^i$ is the minimum input of $u_{pi}$, and $U_{\max}^i$ is the maximum input of $u_{pi}$.

For this control problem, the objective of this study is to propose agents $g_{ci}$ that satisfies the following:

Specification 1: The input of each agent $g_{ci}$ is only the tracking error $e$.

Specification 2: The output $y_p$ of plant tracks the reference signal $y_r$. That is, $y_p \to y_r$ as $t \to \infty$.

Specification 3: The agents $g_{ci}$ have design parameters to adjust their behaviors of tendency in advance, e.g., whether they are easy to stop or not.

## III. Decentralized Control Using Broadcast Signal

This section proposes a distributed control law that satisfies the three specifications given in Section II.B, using only broadcast signals.

Firstly, the error system can be obtained from (1) and (5), as follows:
$$\dot{x} = A_p x + b_p u \\ e = c_p x + d_p u \qquad (6)$$
where $x \triangleq x_r - x_p$, $u \triangleq u_r - u_p$. Then the closed loop system can be represented as shown in Fig. 3. Let the output of $G_{ce}$ be
$$v \triangleq -u. \qquad (7)$$

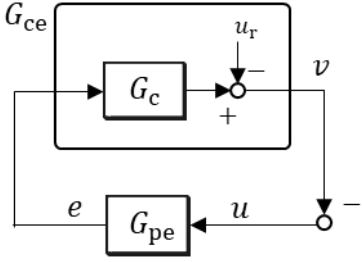

Fig. 3. Error system.

## A. Autonomous Switching Controller (ASC)

Here we refer to the idea of the gait control law described in [30],[31]. The gait control law consists of a mechanism to generate phase differences between legs using nonlinear oscillators to express inter-leg coordination based only on the information about each leg and a switching mechanism to switch between the free and supporting legs, that is, to determine the sections in which control is inactive and active, respectively. In this paper, we exclude the nonlinear oscillator from the mechanism proposed in [30],[31] and propose agents $g_{ci}$ that feeds back only the broadcasted error $e$ (Fig. 4):

$$\dot{\varphi}_i = K_i(\varphi_i, e)e$$
$$u_{pi}(t) = \sigma_i(\varphi_i) = \begin{cases} U_p: \varphi_i > 0 \\ U_n: \varphi_i \leq 0 \end{cases} \quad (8)$$

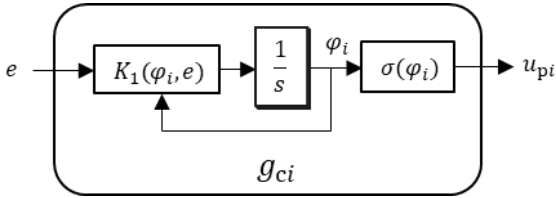

Fig. 4. Autonomous switching controller (ASC).

The control law in (8) is referred to as an Autonomous Switching Controller (ASC). Equation (8) consists of the dynamics that generate the phase $\varphi_i$ of each agent and the switching function that expresses the role assignment function $\sigma_i$. Here $U_p$ and $U_n$ can be freely designed within the range $(U_{\min} \leq mU_n \leq u_r \leq mU_p \leq U_{\max})$ that satisfies Assumption 3. In addition, to design the behavioral tendency of each agent, a variable gain $K_i(\varphi_i, e)$

$$K_i(\varphi_i, e) = \begin{cases} \overline{k}_i(\varphi_i, e) : \varphi_i e < 0 \\ \underline{k}_i(\varphi_i, e) : \varphi_i e \geq 0 \end{cases} \quad (9)$$

is introduced, where $\overline{k}_i(\varphi_i)$ and $\underline{k}_i(\varphi_i)$ are positive functions satisfying $0 < \underline{k}_i(\varphi_i) \leq \overline{k}_i(\varphi_i)$. In the proposed method, the ASC with $U_p$, $U_n$, and $K_i$ is designed such that the role-sharing among the agents is generated based on the characteristics and constraints of the actuators.

For the agents $g_{ci}$ with (8) and (9), the following theorem holds.

Theorem 1
Consider the error system described by equations (6) and (7) using the agents $g_{ci}$ described by equations (3), (8) and (9) and suppose that the system satisfies Assumptions 1, 2 and 3. Then, the tracking error $e$ satisfies $e \to 0$ as $t \to \infty$.
Proof
See Appendix A.1 for the proof. □

When there exists some agents to let stop completely, the $U_n$ is set as 0 in (8). Furthermore, when $e < 0$, it is necessary to immediately set as $u_{pi} = U_n(= 0)$. However, from (8), $u_{pi}(= \sigma(\varphi_i))$ is determined based on the sign of the $\varphi_i$ which is the integral of the error $e$. In other words, there exists a delay until $\varphi_i \leq 0$ becomes true after $e < 0$ becomes true. Therefore, in the proposed controller in (8), the gain $\overline{k}_i$ in (9) is set to be a large value to reduce the delay.

Furthermore, even if some agents failures ($u_{pi} = 0$), the stability is guaranteed by Theorem 1 if Assumption 3 is satisfied.

## B. Autonomous Smooth Switching Controller (ASSC)

Due to the role assignment function $\sigma_i$, the tracking error $e$ become discontinuous and oscillatory. Therefore, to reduce the output oscillation, we propose agents $g_{ci}$ below consisting of a function that continuously interpolates the role assignment function in (8) by $\sigma(\varphi_i)$ as

$$\dot{\varphi}_i = K_i(\varphi_i, e)e$$
$$u_{pi} = \sigma(\varphi_i) = \begin{cases} U_p: \varphi_i \geq \varphi_{pi} \\ \sigma_c(\varphi_i): \varphi_{ni} < \varphi_i < \varphi_{pi} \\ U_n: \varphi_i \leq \varphi_{ni} \end{cases} \quad (10)$$

Here, $\sigma_c(\varphi_i)$ is a non-decreasing function that changes continuously from $U_n$ to $U_p$, taking $\sigma_c(\varphi_{ni}) = U_n(\leq 0)$, $\sigma_c(0) = 0$ and $\sigma_c(\varphi_{pi}) = U_p(> 0)$. The control law in (10) is referred to as an Autonomous Smooth Switching Controller (ASSC). The proposed ASSC method has the design parameters $U_p, U_n, \sigma_c$, and $K_i$.

In the system shown in Fig. 4, where $g_{ci}$ of (10) is used, Theorem 2 holds.

Theorem 2
Consider the error system described by equations (6) and (7) using the agents $g_{ci}$ described by equations (3),(9) and (10) and suppose that the system satisfies Assumptions 1, 2 and 3. Then, the tracking error $e$ satisfies $e \to 0$ as $t \to \infty$.

Proof
See Appendix A.4 for the proof.

## IV. NUMERICAL EXPERIMENTS

This section demonstrates the effectiveness of the proposed method described in Sections III.A and III.B in comparison with the controller in Fig. 5, which is a simplified version of the distributed controller[4] commonly known as an automatic generation control (AGC):

$$\begin{aligned}\dot{\varphi}_i &= K_i e \\ u_{\mathrm{p}i} &= \varphi_i .\end{aligned} \quad (11)$$

Here, the gain $K_i$ is a positive design parameter. Hereafter, the controller in (11) is referred to as an integral controller. See Appendix3 for the stability of the system using (11), $g_{ci}$.

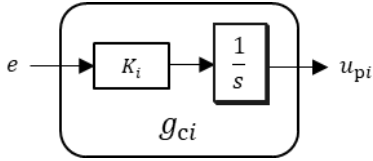

Fig. 5. Decentralized integral control.

### A. Conditions

Let us consider the control problem associated with making the output $y_\mathrm{p}$ of the second-order system given by the following equation and tracking the reference signal $y_\mathrm{r}$, using ten agents ($m = 10$):

$$y_\mathrm{p} = \frac{\beta_1 s + \beta_0}{s^2 + \alpha_1 s + \alpha_0} u_\mathrm{p} \quad (12)$$

where $s$ is the Laplace variable, $\alpha_1 = 98$, $\alpha_0 = 4900$, $\beta_1 = 75$, and $\beta_0 = 4900$. For the control input $u_{\mathrm{p}i}$ in (11), the maximum value is $U_\mathrm{p} = 3$, and the minimum value is $U_\mathrm{n} = 0$.

For the above objects, numerical experiments were conducted under the following conditions.

1) Condition 1 (verification of tracking performance)

The reference signal $y_\mathrm{r}$ was switched stepwise (two-stage). Specifically, $y_\mathrm{r} = 28$ was set for the time interval $0 \leq t < 0.2$ s, and $y_\mathrm{r} = 10$ was set for the time interval $0.2 \leq t < 0.4$ s.

2) Condition 2 (verification of fault-tolerant performance)

In this case, it was supposed that the control inputs $u_{\mathrm{p}i}$ for five agents ($i = 1, ..., 5$) suddenly became zero at $t \geq 0.2$ s owing to some malfunctions. Note that the reference signal $y_\mathrm{r}$ was set as $y_\mathrm{r} = 10$ at all time.

Remark: How to set $y_\mathrm{r}$ as any constant value.

The reference signal $y_\mathrm{r}$ is the output of (5) for a constant input $u_\mathrm{r}$. In other words, using (12), $y_\mathrm{r}$ becomes

$$\begin{aligned}\dot{x}_\mathrm{r} &= \begin{bmatrix} -\alpha_1 & -\alpha_0 \\ 1 & 0 \end{bmatrix} x_\mathrm{r} + \begin{bmatrix} 1 \\ 0 \end{bmatrix} u_\mathrm{r} \\ y_\mathrm{r} &= [\beta_1 \; \beta_0] x_\mathrm{r} .\end{aligned} \quad (13)$$

Therefore, if $y_\mathrm{r}$ is constant, the initial state $x_\mathrm{r}(0)$ can be set as

$$x_\mathrm{r}(0) = -\begin{bmatrix} -\alpha_1 & -\alpha_0 \\ 1 & 0 \end{bmatrix}^{-1} \begin{bmatrix} 1 \\ 0 \end{bmatrix} u_\mathrm{r} = -\begin{bmatrix} 0 \\ 1 \end{bmatrix} u_\mathrm{r} . \quad (14)$$

Then, $\dot{x}_\mathrm{r} = 0$ and $y_\mathrm{r}$ becomes a constant value:

$$y_\mathrm{r} = -[\beta_1 \; \beta_0] \begin{bmatrix} 0 \\ 1 \end{bmatrix} u_\mathrm{r} = -\beta_0 u_\mathrm{r} . \quad (15)$$

### B. Integral Controller Results

The design parameters of the integral controller in (11) are $K_i = 5(2 - 0.2(i - 1))$, and the initial values of the agents are $\varphi_i(0) = 0$ for $i = 1, 2, ..., 10$. We set all the initial values the same to show that their behaviors change even if they are originally identical. The gain $K_i$ with small number is set to be large, that is, the control input is prone to change. To make the comparison conditions the same as those in the proposed method, we imposed the upper and lower limits of control input.

Fig. 6 shows the control results under Condition 1. The control inputs generated by agents 1–5 are represented by solid lines, and those generated by agents 6–10 are represented by dotted lines. In the first half of 0 to 0.2 s of Fig. 6(a), the output $y_\mathrm{p}$ did not track the reference signal $y_\mathrm{r}$. Because some agents had small gains, sufficient input did not be generated even if the error was large. In addition, in Figs. 6(b) and (c), the change patterns of the agents were the same even if the agents had different gains. In other words, the integral controller in (11) with different gains did not generate a clear division of roles.

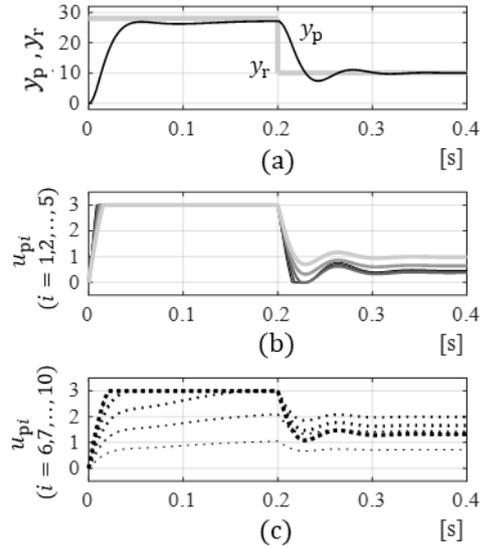

Fig. 6. Integral control results: (a) output $y_\mathrm{p}$ and reference signal $y_\mathrm{r}$; (b, c) control input generated by (b) agents 1–5, (c) agents 6–10.

### C. ASC Results

The design parameters of ASC in (8) and (9) were, for $i = 1,2,...,10$,

$$\underline{k}_i = 5(2 - 0.2(i-1)) \quad : \varphi_i e > 0$$
$$\overline{k}_i = \begin{cases} \underline{k}_i & : \varphi_i \leq 0 \text{ and } e \geq 0 \\ \underline{k}_i(1 + 0.2(i-1)) : \varphi_i > 0 \text{ and } e < 0 \end{cases} \quad (16)$$

Here, the agents with small numbers are prone to be active, that is, they select $U_p$ because of the large values of $\underline{k}_i$. On the other hands, the agents with large numbers are prone to be inactive, that is, they select $U_n$ because of the small values of $\underline{k}_i$ and $\overline{k}_i$ larger than $\underline{k}_i$ for $\varphi_i > 0$ and $e < 0$. Let the initial values of the agents be $\varphi_i(0) = 0$ ($i = 1, 2, ..., 10$).

Figs. 7 and 8 show the control results under Conditions 1 and 2, respectively. Here, the behaviors of agents 1–5 are represented by solid lines, and those of agents 6–10 are represented by dotted lines. In Fig. 7, when the reference signal was set to a large value in the first half of 0 to 0.2 s, the agents selected $U_p(=3)$ and became active. In the region where $y_r < y_p$ after exceeding the target value, agents 9 and 10 selected $U_n(=0)$ and became inactive. After that, agents 1-8 selected $U_p$ and the other agents 9 and 10 switched $U_p$ and $U_n$ alternately. In other words, the agents with small $\underline{k}$ selected not only $U_n$ but also $U_p$ according to the tracking error.

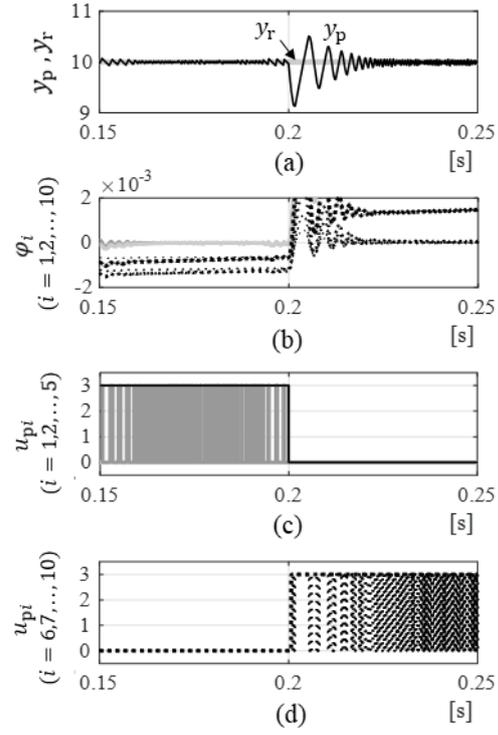

Fig. 8. ASC results (Condition 2): (a) output $y_p$ and target $y_r$, (b) phase $\varphi_i$ of each agent; (c, d) control input generated by (c) agents 1–5, (d) agents 6–10.

In the second half of 0.2 to 0.4 s as shown in Fig.7, the reference signal was set to a small value. In the region where $y_r < y_p$ before reaching the target value, the agents with large number selected $U_n$ became inactive. After that, agents 6-10 stopped in most region because $\varphi_i$ did not increase since $\underline{k}_i$ was small and $\overline{k}_i$ was larger than $\underline{k}_i$ when $\varphi_i > 0$ and $e < 0$. Agents 1-3 selected $U_p$ in most region and agent 4 switched between $U_p$ and $U_n$ alternately and the agents 5 selected $U_n$. In other words, the agents designed to be prone to stop selected $U_n$ and the remaining agents were adjusted to achieve the tracking control. As a result, the roles of agents were automatically assigned without receiving individual commands. In addition, from (9), the output was adjusted to track the reference signal by switching between $U_p$ and $U_n$.

On the other hand, in Condition 2 (Fig. 8), even when agents 1–5 broke down at 0.2 s ($u_{pi} = 0$), agent 6 automatically selected $U_p$ and the other agents 7-10 switched between $U_p$ and $U_n$ (=0) to track to the reference.

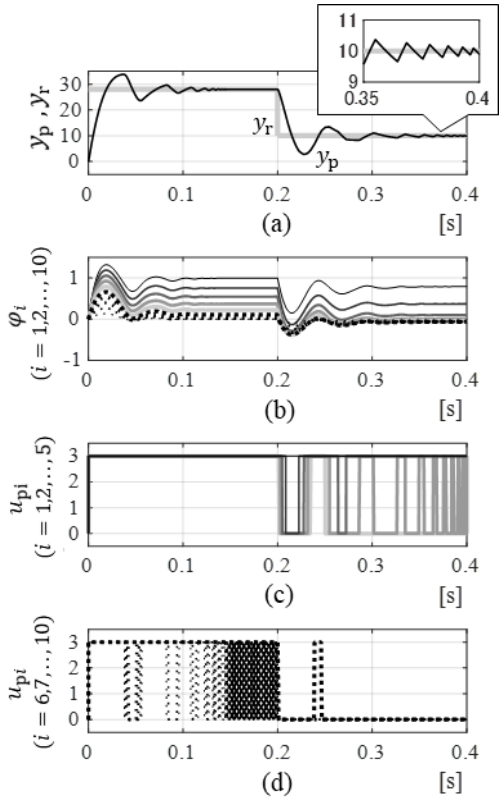

Fig. 7. ASC results (Condition 1): (a) output $y_p$ and reference signal $y_r$, (b) phase $\varphi_i$ of each agent; (c, d) control input generated by (c) agents 1–5, (d) agents 6–10.

### D. ASSC Results

The design parameters $\overline{k}_i$, $\underline{k}_i$ of ASSC in (10) were set to (16) as well. Let the initial values of the agents be $\varphi_i(0) = 0$ ($i = 1, 2, ..., 10$). Furthermore, the switching function was

$$\sigma_c(\varphi_i) \triangleq \frac{U_p}{\varphi_{pi} - \varphi_{ni}}(\varphi_i - \varphi_{ni}) \quad (17)$$
$$\varphi_{ni} = 0, \quad \varphi_{pi} = 0.06.$$

The switching function connected $U_p$ and $U_n$ with a straight line. In addition, $\varphi_{pi}$ was set to a small value to clarify the division, that is, to select either $U_p$ or $U_n$.

Fig. 9 shows the control results under Condition 1. In Fig. 9, behaviors of agents 1–5 and 6–10 are represented by solid and dotted lines, respectively. Comparing Figs. 7(a) and 9(a), it was apparently that the output oscillations were suppressed. This suppression occurs because, as shown in Figs. 9(c) and (d), there were agents that selected intermediate values other than $U_p$ and $U_n$. Between 0 and 0.2 s, when the reference signal was large, agents 1-7 selected $U_p$, agents 8-10 selected intermediate values or $U_p$. Meanwhile, when the reference signal was small between 0.2 and 0.4 s, agents 6-10 selected $U_n$ and agents 1-5 selected intermediate values from $U_p$ to $U_n$ in transient state. In steady state, agent 1-3 selected $U_p$, agent 4 selected intermediate values and agent 5 selected $U_n$. Therefore, the division of roles based on the predesigned behavioral tendencies was expressed, and tracking control was achieved while reducing the oscillation.

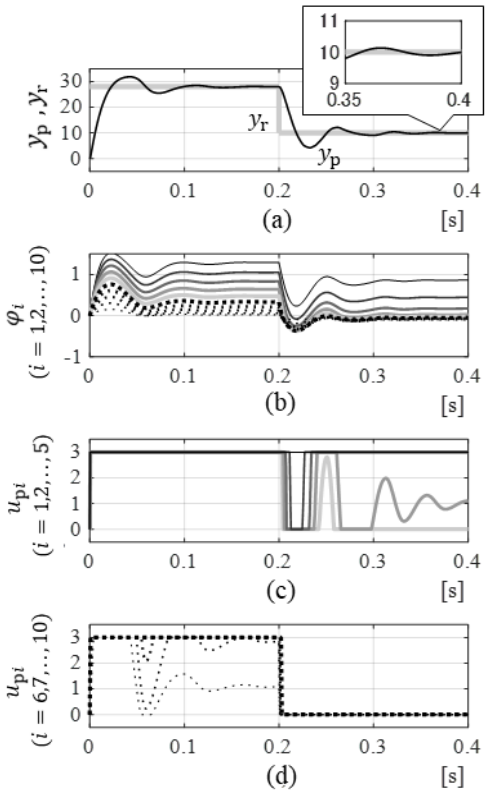

Fig. 9. Autonomous smooth switching controller (ASSC) results (Condition 1): (a) output $y_p$ and target $y_r$, (b) phase $\varphi_i$ of each agent; (c, d) control input generated by (c) agents 1–5, (d) agents 6–10.

## V. DISCUSSION

The multi-agent system $G_c$ proposed in this paper achieves the tracking control by autonomously assigning roles to different agents according to the magnitude of the reference signal or the fault such that causes some agents to be zero. In addition, by changing the control parameters, the behavioral tendencies of the agents such as active or inactive are designed in advance. More specifically, the behavior of the agent, which is determined based on the broadcast signal $e$ only, is changed by adjusting the dynamic response of the variable $\varphi_i$ by setting the gain $K_i$.

Conventional broadcast control is an effective method for a system composed of many agents but involves a probabilistic search process. Therefore, it is difficult to control the behavior of each agent. As mentioned above, the proposed method allows the movement of each agent or group of agents to be specified, which makes it easier to manage the agents.

Furthermore, by the smooth switching controller, oscillations of outputs in the steady state are suppressed without impairing the automatically division of roles such as active or inactive.

## VI. CONCLUSION

In this paper, we proposed a deterministic controller for a multi-agent system that tracks a reference signal using the sum of the control inputs from several agents, and proved that the tracking error is asymptotically stable. Further, we realized a broadcast controller without introducing any stochastic mechanism to enable the multi-agent system to generate the division of roles.

The proposed methods enable the construction of efficient and reliable systems. For a multi-motor system, the failure of one motor can be automatically compensated by the other motors, and the motors can be turned on according to their efficiencies. For an energy management system with multiple power supply sources, the shortage power can be compensated by starting dormant sources automatically, and highly efficient sources can be used preferentially.

Nevertheless, the limitation of the proposed method is that the asymptotic stability is guaranteed for strictly positive real systems. In the future, we plan to relax this constraint and make it applicable to more general systems.

## APPENDIX

### A.1 Proof of Theorem 1

In the system shown in Fig. 3, if $G_{\text{pe}}$ is strictly positive real and $G_{\text{ce}}$ is passive, the hyperstability theorem [32],[33] guarantees the asymptotic stability of the error $e$. Because $G_{\text{pe}}$ is assumed to be strictly positive real, it is sufficient to show that $G_{\text{ce}}$ is passive.

Firstly, by introducing $u_{ri} \in R$ ($i = 1, 2, \dots, m$) which is the share of each agent for the reference input

$$u_r \triangleq \sum_{i=1}^{m} u_{ri}. \tag{A.1}$$

Here, because $u_{ri}$ is the share of each agent, the value is always within the range between the maximum and minimum control inputs, that is, $U_n \leq u_{ri} \leq U_p$.

Using this $u_{ri}$ the output of $G_{ce}$ can be expressed as follows:

$$v = \sum_{i=1}^{m} v_i \qquad v_i \triangleq u_{pi} - u_{ri} \tag{A.2}$$

where $u_{pi} = \sigma(\varphi_i)$.

Regarding (8) and (9), the storage function $V_c$ for the overall $G_{ce}$ and the storage function $V_{ci}$ of each agent can be defined as follows:

$$V_c \triangleq \sum_{i=1}^{m} V_{ci} \qquad V_{ci} \triangleq \int_{0}^{\varphi_i} \frac{\sigma(\varphi_i) - u_{ri}}{L_i(\varphi_i)} d\varphi_i \geq 0 \tag{A.3}$$

where $L_i$ is an arbitrary positive function satisfying $\underline{k_i}(\varphi_i) \leq L_i(\varphi_i) \leq \overline{k_i}(\varphi_i)$. The derivatives of $V_{ci}$ are

$$\dot{V}_{ci} = \frac{\sigma(\varphi_i) - u_{ri}}{L_i(\varphi_i)} \dot{\varphi}_i = \frac{K_i(\varphi_i, e)}{L_i(\varphi_i)} (\sigma(\varphi_i) - u_{ri}) e. \tag{A.4}$$

Accordingly, using (A.3), the derivative of $V_c$ can be expressed as follows:

$$\dot{V}_c = \sum_{i=1}^{m} \dot{V}_{ci} = \sum_{i=1}^{m} \frac{K_i(\varphi_i, e)}{L_i(\varphi_i)} (\sigma(\varphi_i) - u_{ri}) e$$

$$= \sum_{i=1}^{m} \frac{K_i(\varphi_i, e)}{L_i(\varphi_i)} v_i e$$

$$= ve + \sum_{i=1}^{m} \left( \frac{K_i(\varphi_i, e)}{L_i(\varphi_i)} - 1 \right) v_i e. \tag{A.5}$$

Using (9), as $\underline{k_i}(\varphi_i) \leq L_i(\varphi_i) \leq \overline{k_i}(\varphi_i)$,

$$\frac{K_i(\varphi_i, e)}{L_i(\varphi_i)} = \begin{cases} \geq 1: \varphi_i e < 0 \\ \leq 1: \varphi_i e \geq 0 \end{cases} \tag{A.6}$$

holds. Moreover, according to the definition, because the signs of $\varphi_i$ and $v_i$ match, $\varphi_i e < 0$ for $v_i e < 0$. Using (A.6), we obtain

$$\left( \frac{K_i(\varphi_i, e)}{L_i(\varphi_i)} - 1 \right) v_i e \leq 0. \tag{A.7}$$

In addition, because $\varphi_i e \geq 0$ for $v_i e \geq 0$, using (A.6), we obtain

$$\left( \frac{K_i(\varphi_i, e)}{L_i(\varphi_i)} - 1 \right) v_i e \leq 0. \tag{A.8}$$

Accordingly, the following is true:

$$\dot{V}_c \leq ve \tag{A.9}$$

showing that $G_{ce}$ is passive. Therefore, the asymptotic stability of error $e$, that is, $e \to 0$ as $t \to \infty$, is proven.
□

### A.2 Proof of Theorem 2

As in Theorem 1, the multi-agent system $G_c$ is passive and the storage function of each agent is given by

$$V_{ci} \triangleq \int_{0}^{\varphi_i} \frac{\sigma(\varphi_i) - u_{ri}}{L_i(\varphi_i)} d\varphi_i \geq 0. \tag{A.10}$$

In the case of (8), as shown in Fig. 1(a), if $\varphi_i \geq 0$, then

$$\sigma(\varphi_i) - u_{ri} = U_p - u_{ri} \geq 0 \tag{A.11}$$

and if $\varphi_i < 0$, then

$$-\sigma(\varphi_i) + u_{ri} = -U_n + u_{ri} \geq 0. \tag{A.12}$$

However, when $\sigma$ is made continuous (smooth), as shown in Fig. 1(b), if $\varphi_i \geq \varphi_{pi}$, then

$$\sigma(\varphi_i) - u_{ri} = U_p - u_{ri} \geq 0 \tag{A.13}$$

holds, but if $\varphi_i < \varphi_{pi}$, then

$$\sigma(\varphi_i) - u_{ri} \geq 0 \tag{A.14}$$

is not necessarily satisfied. Similarly, as shown in Fig. A.1(b), if $\varphi_i \leq \varphi_{ni}$, then

$$-\sigma(\varphi_i) + u_{ri} = -U_n + u_{ri} \geq 0 \tag{A.15}$$

holds, but if $\varphi_i > \varphi_{ni}$, then

$$-\sigma(\varphi_i) + u_{ri} \geq 0 \tag{A.16}$$

is not necessarily satisfied. Accordingly, in (A.10), it can not be said that $V_{ci} \geq 0$ across the entire region, making it unsuitable for use as a storage function. With respect to (A.10), a new variable $\tilde{u}_{ri}$ is introduced, and the storage function $V_{ci}$ is given a new definition:

$$V_{ci} \triangleq \int_{0}^{\varphi_i} \frac{\sigma(\varphi_i) - \tilde{u}_{ri}(\varphi_i)}{L_i(\varphi_i)} d\varphi_i \geq 0. \tag{A.17}$$

$\tilde{u}_{ri}(\varphi_i)$ in (A.17) is a function, as shown in Fig. A.2, where for $u_{ri} > 0$,

$$\tilde{u}_{ri}(\varphi_i) = \begin{cases} u_{ri} : \varphi_i \geq \sigma^{-1}(u_{ri}) \\ \sigma(\varphi_i) : \varphi_i < \sigma^{-1}(u_{ri}) \end{cases} \tag{A.18}$$

and for $u_{ri} \leq 0$,

$$\tilde{u}_{ri}(\varphi_i) = \begin{cases} u_{ri} : \varphi_i \leq \sigma^{-1}(u_{ri}) \\ \sigma(\varphi_i) : \varphi_i > \sigma^{-1}(u_{ri}) \end{cases} \tag{A.19}$$

are defined.

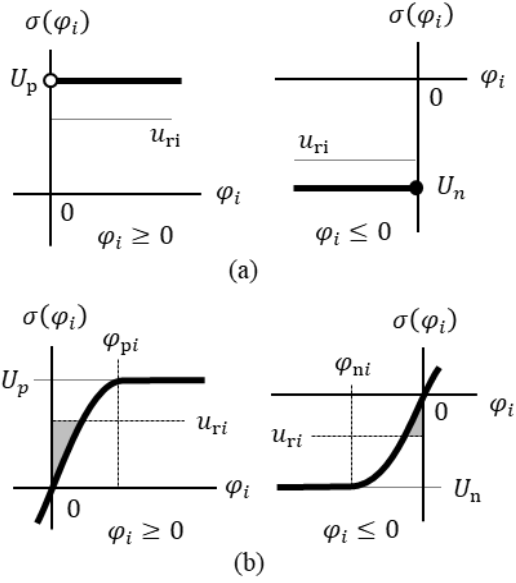

Fig. A.1 (a) Two-stage and (b) smooth switching σ.

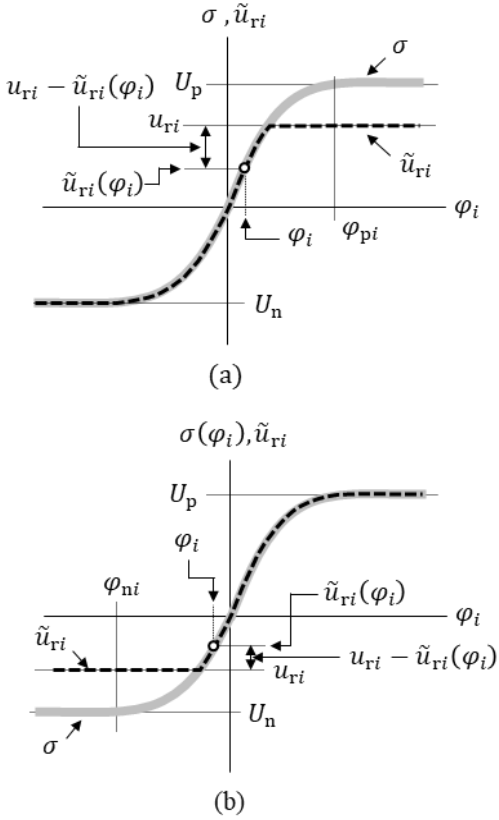

Fig. A.2 Definition of $\tilde{u}_{ri}$ for (a) $u_{ri} > 0$ and (b) $u_{ri} \leq 0$.

Based on the above definitions, the derivative of the storage function $V_{ci}$ in (A.17) can be expressed as follows:

$$\begin{aligned}\dot{V}_{ci} &= \frac{\sigma(\varphi_i) - \tilde{u}_{ri}(\varphi_i)}{L_i(\varphi_i)}\dot{\varphi}_i = KL_i\big(\sigma(\varphi_i) - \tilde{u}_{ri}(\varphi_i)\big)e \\ &= KL_i(\sigma(\varphi_i) - u_{ri})e + KL_i\big(u_{ri} - \tilde{u}_{ri}(\varphi_i)\big)e \\ &= (\sigma(\varphi_i) - u_{ri})e(t) + (KL_i - 1)(\sigma(\varphi_i) - u_{ri})e \\ &\quad + KL_i\big(u_{ri} - \tilde{u}_{ri}(\varphi_i)\big)e \\ &= (\sigma(\varphi_i) - u_{ri})e + (KL_i - 1)(\sigma(\varphi_i) - \tilde{u}_{ri}(\varphi_i) \\ &\quad - (u_{ri} - \tilde{u}_{ri}(\varphi_i))e + KL_i\big(u_{ri} - \tilde{u}_{ri}(\varphi_i)\big)e \\ &= v_i e + (KL_i - 1)\big(\sigma(\varphi_i) - \tilde{u}_{ri}(\varphi_i)\big)e \\ &\quad + \big(u_{ri} - \tilde{u}_{ri}(\varphi_i)\big)e \end{aligned} \quad (A.20)$$

where $KL_i \triangleq K_i(\varphi_i, e)/L_i(\varphi_i)$.

Next, $\tilde{u}_r$ is defined as follows:

$$\tilde{u}_r \triangleq \sum_{i=1}^{m} \tilde{u}_{ri}(\varphi_i). \quad (A.21)$$

With regard to (10), if the storage function for the overall $G_{ce}$ is defined as

$$V_c \triangleq \sum_{i=1}^{m} V_{ci} \quad (A.22)$$

then, using (A.20) and (A.21), the derivative of $V_{ci}$ can be expressed as

$$\dot{V}_c = ve + \sum_{i=1}^{m}\big(u_{ri} - \tilde{u}_{ri}(\varphi_i)\big)e$$

$$+ \sum_{i=1}^{m}(KL_i - 1)\big(\sigma(\varphi_i) - \tilde{u}_{ri}(\varphi_i)\big)e. \quad (A.23)$$

Here, we define a new function $V_{ui}$ as

$$V_{ui} \triangleq \int_0^{\varphi_i}\big(u_{ri} - \tilde{u}_{ri}(\varphi_i)\big)\,d\varphi_i \geq 0. \quad (A.24)$$

The derivative function $\dot{V}_{ui}$ is

$$\dot{V}_{ui} = \big(u_{ri} - \tilde{u}_{ri}(\varphi_i)\big)\dot{\varphi}_i = K_i(\varphi_i, e)\big(u_{ri} - \tilde{u}_{ri}(\varphi_i)\big)e. \quad (A.25)$$

Substituting the above equation into the second term on the right-hand side of (A.23), we obtain

$$\sum_{i=1}^{m}\big(u_{ri} - \tilde{u}_{ri}(\varphi_i)\big)e = \sum_{i=1}^{m}\left(\frac{1}{K_i(\varphi_i, e)}\right)\dot{V}_{ui}. \quad (A.26)$$

Next, integrating both sides of (A.23) over time interval $[t_0\ t_1]$ provides

$$V_c(t_1) - V_c(t_0) = \\ \int_{t_0}^{t_1} ve\,dt + \int_{t_0}^{t_1}\sum_{i=1}^{m}(KL_i - 1)\big(\sigma(\varphi_i) - \tilde{u}_{ri}(\varphi_i)\big)e\,dt \\ + \int_{t_0}^{t_1}\sum_{i=1}^{m}\big(u_{ri} - \tilde{u}_{ri}(\varphi_i)\big)e\,dt. \quad (A.27)$$

Based on the definitions of $K_i(\varphi_i, e)$ in (10) and $L_i(\varphi_i)$ in (A.6),

$$\begin{aligned} \varphi_i e \leq 0 &\to (KL_i - 1) \geq 0 \\ \varphi_i e \geq 0 &\to (KL_i - 1) \leq 0 \end{aligned} \quad (A.28)$$

Furthermore, by the definition of $\tilde{u}_{ri}(\varphi_i)$,

1) $\varphi_i e \leq 0$

$$\begin{aligned} \varphi_i > 0, e \leq 0 &\to (\sigma(\varphi_i) - \tilde{u}_{ri})e \leq 0 \\ \varphi_i \leq 0, e > 0 &\to (\sigma(\varphi_i) - \tilde{u}_{ri})e \leq 0 \end{aligned} \quad (A.29)$$

2) $\varphi_i e \geq 0$

$$\begin{aligned} \varphi_i > 0, e > 0 &\to (\sigma(\varphi_i) - \tilde{u}_{ri})e \geq 0 \\ \varphi_i \leq 0, e \leq 0 &\to (\sigma(\varphi_i) - \tilde{u}_{ri})e \geq 0 \,. \end{aligned} \quad (A.30)$$

Hence, the second term on the right-hand side of (A.27) is non-positive.

Substituting (A.26) into the third term on the right-hand side of (A.27), we obtain

$$\int_{t_0}^{t_1} \sum_{i=1}^{m} (u_{ri} - \tilde{u}_{ri}(\varphi_i))e \, dt = \sum_{i=1}^{m} \int_{t_0}^{t_1} \left( \frac{1}{K_i(\varphi_i, e)} \dot{V}_{ui} \right) dt \quad (A.31)$$

where we introduce $\dot{\bar{V}}_{ui}$, which is the upper limit of $\dot{V}_{ui}$

$$\dot{\bar{V}}_{ui} \triangleq \begin{cases} \dot{V}_{ui} : \dot{V}_{ui} \geq 0 \\ 0 : \dot{V}_{ui} < 0 \,. \end{cases} \quad (A.32)$$

This changes (A.31) to

$$\int_{t_0}^{t_1} \sum_{i=1}^{m} (u_{ri} - \tilde{u}_{ri}(\varphi_i))e \, dt \leq \sum_{i=1}^{m} kh \int_{t_0}^{t_1} \dot{\bar{V}}_{ui} \, dt \,. \quad (A.33)$$

Here, because $K_i(\varphi_i, e)$ is a positive design parameter, there exists a positive constant $kh$, which is the upper limit of $0 < 1/K_i(\varphi_i, e) \leq kh$. Now, let the initial time be $tt_0 = t_0$, the end time be $tt_n = t_1$, and the interval in which $\dot{V}_{ui}$ is a positive or non-positive term be $tt_{i-1} \sim tt_i$ ($i = 1, 2, \ldots, n$), where $n$ is the total number of intervals in which $\dot{V}_{ui}$ is positive and non-positive.

Using the interval, the right-hand side of (A.33) becomes

$$\sum_{i=1}^{m} kh \int_{t_0}^{t_1} \dot{\bar{V}}_{ui} dt$$

$$= \sum_{i=1}^{m} kh(V_{ui}(tt_1) - V_{ui}(tt_0) + V_{ui}(tt_2) - V_{ui}(tt_1) +$$

$$\cdots + V_{ui}(tt_n) - V_{ui}(tt_{n-1}))$$

$$= kh \sum_{i=1}^{m} (V_{ui}(tt_n) - V_{ui}(tt_0)). \quad (A.34)$$

At this point, we have $u_{ri} - r\tilde{u}_{ri}(\varphi_i) = 0$ when $\varphi_i \geq \varphi_{pi}$ or $\varphi_i < \varphi_{ni}$, so that $V_{ui}$ is

$$V_{ui} = \int_0^{\varphi_i} (u_{ri} - \tilde{u}_{ri}(\varphi_i)) \, d\varphi_i$$

$$\leq \int_0^{\varphi_i} \Delta u_{rm} d\varphi_i = \Delta u_{rm} |\varphi_i| \leq \Delta u_{rm} \varphi_{mi} \,. \quad (A.35)$$

Here, $\Delta u_{rm} \triangleq \max_{\varphi_i} \sigma(\varphi_i) - \min_{\varphi_i} \sigma(\varphi_i)$ and $\varphi_{mi} \triangleq \max\{|\varphi_{pi}|, |\varphi_{ni}|\}$. Therefore, (A.34) is

$$kh \sum_{i=1}^{m} (V_{ui}(tt_n) - V_{ui}(tt_0)) \leq kh \, m \, \Delta u_{rm} \varphi_{mi} \quad (A.36)$$

Thus,

$$\int_{t_0}^{t_1} \sum_{i=1}^{m} (u_{ri} - \tilde{u}_{ri}(\varphi_i))e \, dt \leq C_u \quad (A.37)$$

where $0 \leq C_u \triangleq m \, kh \, \Delta u_r \varphi_{mi} < \infty$.
Hence, (A.27) is

$$V_c(t_1) - V_c(t_0) \leq \int_{t_0}^{t_1} ve \, dt + C_u \,. \quad (A.38)$$

As $CV_c(t_1) \geq 0$ and $C_u \geq 0$,

$$-C_{Vu}(t_0) \leq \int_{t_0}^{t_1} ve \, dt \,. \quad (A.39)$$

Here, $0 \leq C_{Vu}(t_0) \triangleq V_c(t_0) + C_u < \infty$. Hence, the passivity of $G_c$ has been proven. As a result, the asymptotic stability of the error is compensated even for agent $g_{ci}$ in (10).

### A.3. Stability of the Integral Controller System

This section demonstrates that stability can be guaranteed even by using the integral controller of (11).

Theorem 3
Applying $g_{ci}$ given by (3) and (11) to the error system defined by (6) and (7) and satisfying Assumptions 1 and 2, the system is asymptotically stable and $e \to 0$ as $t \to \infty$.

Proof
As in Theorem 1, we show that the multi-agent system $G_c$ is passive. We define the storage function $V_c$ of $G_{ce}$ in (11) as

$$V_c \triangleq \frac{1}{2K_s} \left( \sum_{i=1}^{m} u_{pi} - u_r \right)^2 . \quad (A.40)$$

Here, $K_s \triangleq \sum_{i=1}^{m} K_i$ is a constant. Because $u_r$ is a constant and from (3) and (7), the derivative of the storage function $V_c$ is

$$\dot{V}_c \triangleq \frac{1}{K_s} \left( \sum_{i=1}^{m} u_{pi} - u_r \right) \left( \sum_{i=1}^{m} \dot{u}_{pi} \right)$$

$$= \frac{1}{K_s} (u_p - u_r) \left( \sum_{i=1}^{m} K_i \right) e = ve \quad (A.41)$$

which demonstrates the passivity of $G_{ce}$. Thus, the asymptotic stability of the error is compensated for agent $g_{ci}$ in (11). □

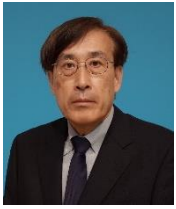
**Yasuhi Amano** received the B.E., M.E., and D.E. degrees in mechanical engineering from Sophia University, Tokyo, Japan, in 1980, 1982, and 1985, respectively. Since 1985, he has been a Researcher with Toyota Central R&D Labs., Inc., Aichi, Japan. His research interests include nonlinear and distributed control theory, design of vehicle control systems, and man–machine systems.

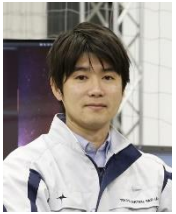
**Tomohiko Jimbo** received the B.E. degree in electronic mechanical engineering, M.E. degree in microsystem engineering, and D.E. degree in mechanical engineering science from Nagoya University, Nagoya, Japan, in 2000, 2002, and 2010, respectively. Since 2002, he has been a Researcher with Toyota Central R&D Labs., Inc., Aichi, Japan. His research interests include modeling, control theory, signal processing, and the system design of automotive engines, power trains, and vehicle/robotic systems

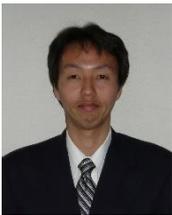
**Kenji Fujimoto** received the B.Sc. and M.Sc. degrees in engineering and Ph.D. degree in Informatics from Kyoto University, Japan, in 1994, 1996, and 2001, respectively. He is currently a Professor with Graduate School of Engineering, Kyoto University, Japan. From 1997 to 2004 he was a Research Associate with Graduate School of Engineering and Graduate School of Informatics, Kyoto University, Japan. From 2004 to 2012 he was an Associate Professor with Graduate School of Engineering, Nagoya University, Japan. He has held visiting research positions at Australian National University, Delft University of Technology, and Groningen University. He received the IFAC Congress Young Author Prize in IFAC World Congress 2005. His research interests include nonlinear control and stochastic systems theory.